\documentclass{article}
\usepackage{ijcai25}
\usepackage{amssymb}
\usepackage{multirow}
\usepackage{color}
\usepackage[table,xcdraw]{xcolor}
\usepackage{graphicx}
\usepackage{booktabs}  %
\usepackage{xcolor} 
\usepackage{array} 

\usepackage{multirow}
\usepackage{times}
\usepackage{soul}
\usepackage{url}
\usepackage[hidelinks]{hyperref}
\usepackage[utf8]{inputenc}
\usepackage[small]{caption}
\usepackage{graphicx}
\usepackage{amsmath}
\usepackage{amsthm}
\usepackage{booktabs}
\usepackage{algorithm}
\usepackage{algorithmic}
\usepackage[switch]{lineno}

\title{PedDet: Adaptive Spectral Optimization for Multimodal Pedestrian Detection}

\author{
Rui Zhao$^{1}$\thanks{Equal contribution. $^\dag$Project lead. $^\ddag$Corresponding author (y.zhao2@latrobe.edu.au).},~ 
Zeyu Zhang$^{23*\dag}$,~ %
Yi Xu$^4$,~ %
Yi Yao$^5$,~ %
Yan Huang$^6$,\\
Wenxin Zhang$^7$,~
Zirui Song$^{68}$,~
Xiuying Chen$^8$,~
Yang Zhao$^{3\ddag}$\\
\affiliations
$^1$JD.com~
$^2$The Australian National University~
$^3$La Trobe University\\
$^4$Central South University~
$^5$NavInfo Co., Ltd.~
$^6$University of Technology Sydney~\\
$^7$University of Chinese Academy of Science~
$^8$Mohamed bin Zayed University of Artificial Intelligence
}

\begin{document}

\maketitle

\begin{abstract}
Pedestrian detection in intelligent transportation systems has made significant progress but faces two critical challenges: (1) insufficient fusion of complementary information between visible and infrared spectra, particularly in complex scenarios, and (2) sensitivity to illumination changes, such as low-light or overexposed conditions, leading to degraded performance. To address these issues, we propose \textbf{PedDet}, an adaptive spectral optimization complementarity framework which specifically enhanced and optimized for multispectral pedestrian detection. PedDet introduces the Multi-scale Spectral Feature Perception Module \textbf{(MSFPM)} to adaptively fuse visible and infrared features, enhancing robustness and flexibility in feature extraction. Additionally, the Illumination Robustness Feature Decoupling Module \textbf{(IRFDM)} improves detection stability under varying lighting by decoupling pedestrian and background features. We further design a contrastive alignment to enhance intermodal feature discrimination. Experiments on \textbf{LLVIP} and \textbf{MSDS }datasets demonstrate that PedDet achieves state-of-the-art performance, improving the mAP by 6.6 \% with superior detection accuracy even in low-light conditions, marking a significant step forward for road safety. 
Code will be available at \url{https://github.com/AIGeeksGroup/PedDet}.
\end{abstract}

\section{Introduction}
Pedestrian detection, a critical component in ensuring road safety, has garnered significant attention from both academia and industry. With the advancement of intelligent transportation systems, pedestrian detection plays an essential role in scenarios such as autonomous driving and public transportation monitoring. However, the complexity of road environments presents significant challenges to pedestrian detection, including issues like occlusion, diverse target sizes, and varying illumination conditions. Addressing these challenges necessitates more precise and robust detection algorithms to enhance traffic safety.

\begin{figure}[t]
	\centering
	\includegraphics[width=0.99\linewidth]{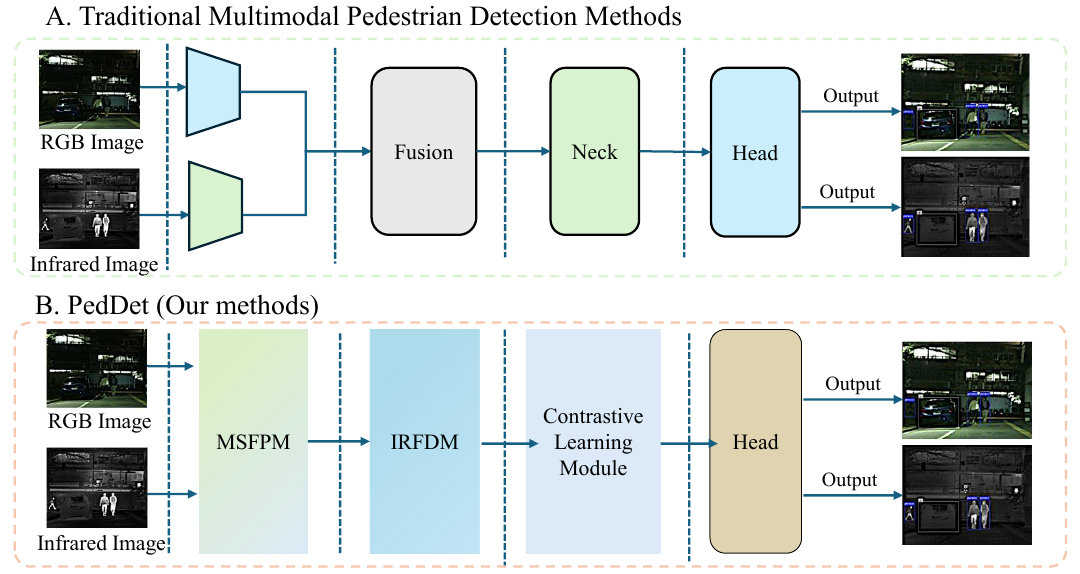}
	\caption{Modules A and B illustrate the distinctions between the existing pedestrian detection frameworks and our proposed approach. Methods in Module A identify and locate targets only by fusing features from multiple levels. Module B shows that our method takes advantage of the Multi-scale Spectral Feature Perception Module (MSFPM) and the Illumination Robustness Feature Decoupling Module (IRFDM) to improve detection performance}
	\label{fig:first}
\end{figure}

\begin{figure*}[t]
	\centering
	\includegraphics[width=\textwidth]{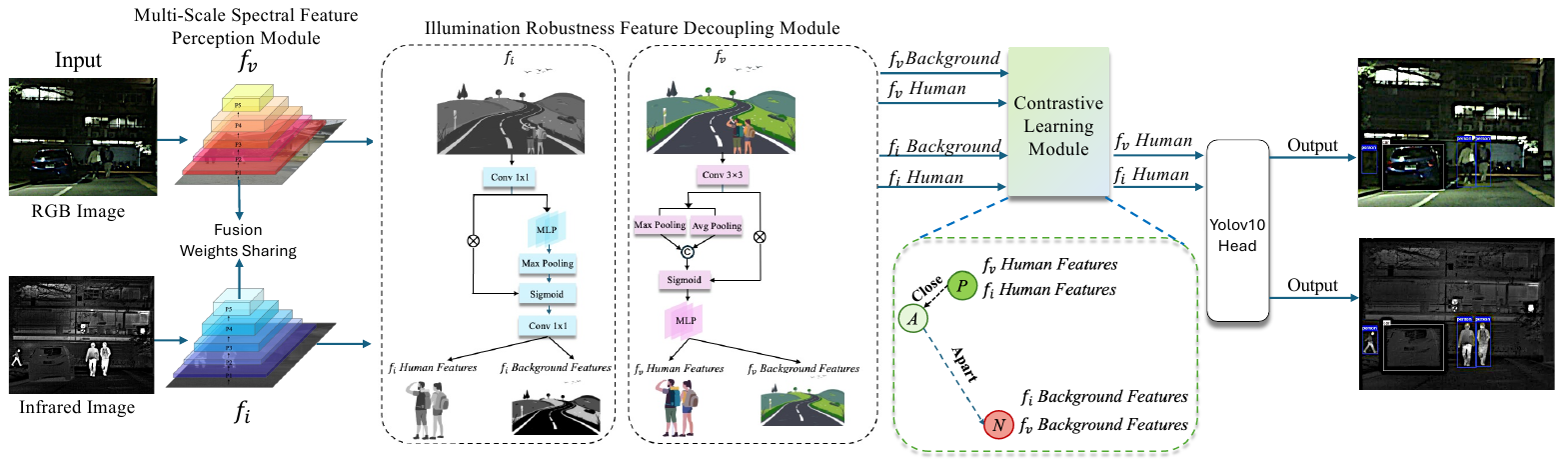}
	\caption{The overall pipeline of PedDet.}
	\label{fig:main}
\end{figure*}

Despite progress in pedestrian detection, current methods face two major challenges. First, existing algorithms struggle to fully exploit the complementary information between RGB and infrared spectra, resulting in suboptimal multimodal feature fusion. In complex scenarios, single-modal pedestrian detection often exhibits instability under unfavorable illumination conditions. Second, traditional methods \cite{1,2,3} demonstrate a high sensitivity to environmental changes, particularly under low-light or overexposed conditions. This sensitivity hampers the models' ability to extract reliable pedestrian features, leading to a significant decline in detection performance. These challenges hinder the widespread real-world deployment of existing pedestrian detection models.

To address these challenges, we propose PedDet, an optimization framework tailored for all object detection models. 
PedDet is specifically designed to optimize pedestrian detection by leveraging the complementary information between RGB and infrared, thereby improving both robustness and accuracy in diverse lighting scenarios, as shown in Figure \ref{fig:first}. 

First, we introduce the Multi-scale Spectral Feature Perception Module (MSFPM), which processes RGB and infrared spectral data in parallel. MSFPM adaptively extracts and fuses features from multiple spectral scales, dynamically adjusting the feature weights of each modality based on illumination conditions. This enables the model to maximize spectral complementarity and maintain robustness across varying environmental complexities. In contrast to traditional single-modal approaches, MSFPM effectively integrates information from both spectra, significantly enhancing detection stability and accuracy.

Additionally, we design the Illumination Robustness Feature Decoupling Module (IRFDM) to address the adverse effects of extreme lighting conditions on detection performance. IRFDM focuses on decoupling pedestrian-specific features from background noise, thereby mitigating interference from complex environments. Traditional pedestrian detection methods often struggle to differentiate pedestrians from the background under extreme conditions, such as strong light, low light, or uneven illumination, where the accuracy significantly deteriorates due to occlusion, shadows, or light reflections. By incorporating IRFDM, our model effectively separates illumination-induced disturbances during feature extraction, ensuring high detection accuracy even under extreme lighting conditions. IRFDM learns robust feature representations across varying illumination scenarios, reducing the negative impacts of environmental lighting variations.

To further improve feature discrimination, we adopt a contrastive learning paradigm. This strategy compares the differences between pedestrian and background features across visible and infrared spectra, strengthening the discrimination between pedestrian and background features and improving the model’s performance in complex scenarios. Comprehensive experiments demonstrate that PedDet achieves state-of-the-art performance, significantly outperforming existing methods in accuracy and robustness under diverse and challenging conditions.

The primary contributions of this work are summarized as follows:
\begin{itemize}
	\item We propose \textbf{PedDet}, which uses a Multi-scale Spectral Feature Perception Module (\textbf{MSFPM}) to fuse visible and infrared features, adjusting weights based on lighting conditions for improved pedestrian detection and accuracy compared to single-modal methods.
	\item To address illumination challenges, we design the Illumination Robustness Feature Decoupling Module (\textbf{IRFDM}), which isolates pedestrian features from background noise, enhancing robustness across different lighting conditions.
	\item A contrastive learning strategy is integrated to improve feature discrimination, helping the model better distinguish pedestrians from background in both visible and infrared modalities, further enhancing robustness in complex environments.
\end{itemize}

\section{Related Works}
Pedestrian detection in intelligent transportation systems has made some strides, but existing studies continue to face two primary challenges: \textit{multimodal feature fusion} and \textit{illumination robustness}. While various methods have been proposed to address these issues, they still exist limitations in real-world scenarios. For related works, see \textbf{Supplementary Section 1: Related Works}.

\section{Methodology}
\subsection{Overview}
In this section, we present the proposed PedDet in detail. As illustrated in Figure \ref{fig:main}, PedDet consists of four key components: an MSFPM, an IRFDM, a contrastive learning paradigm, and a detection head. The following subsections provide an in-depth explanation of each component, concluding with a description of the model’s optimization objectives.

\subsection{Feature Extractor}
\begin{figure}
	\centering
	\includegraphics[width=\linewidth]{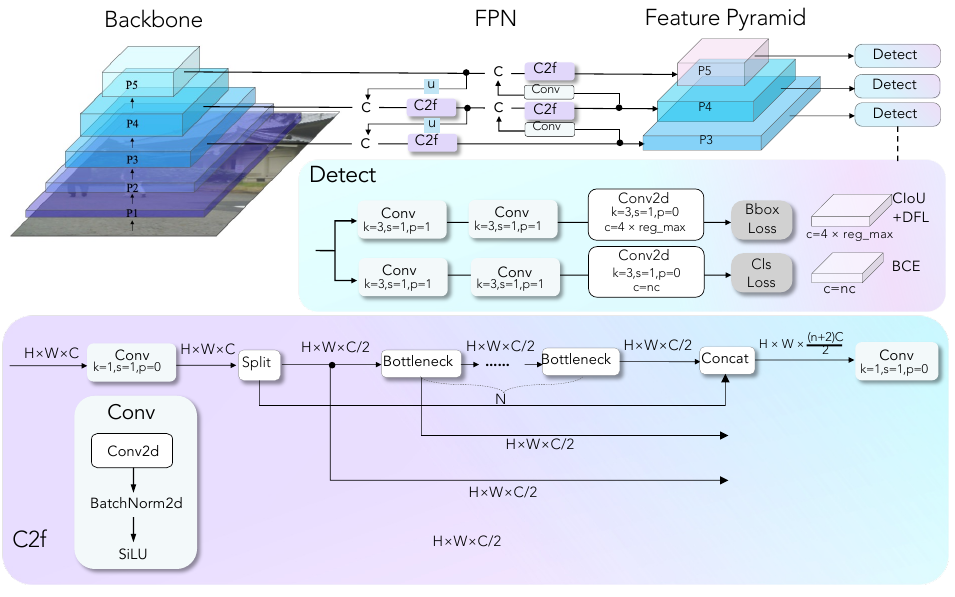}
	\caption{The Architecture of Feature Extractor.}
	\label{fig:extactor}
\end{figure}

We adopt the improved YOLOv10 \cite{4} as the backbone network, as illustrated in Figure \ref{fig:extactor}. YOLOv10 introduces systematic optimizations at multiple levels, significantly enhancing computational efficiency and feature extraction capability. By refining the backbone architecture, YOLOv10 reduces computational complexity and memory consumption while maintaining high model accuracy, making it particularly suitable for real-time applications with stringent latency requirements.

YOLOv10 serves as the backbone for feature extraction, processing RGB and infrared images through parallel branches, each generating modality-specific feature matrices. The RGB branch captures rich color details in daylight, while the infrared branch enhances object contours in low-light conditions. Feature extraction operates across multiple scales, producing feature maps at:  

\begin{equation}
F_1 \in \mathbb{R}^{80 \times 80 \times 256}, \quad  
F_2 \in \mathbb{R}^{40 \times 40 \times 512}, \quad  
F_3 \in \mathbb{R}^{20 \times 20 \times 512}
\end{equation}

where \( F_i \) represents feature maps at different scales. This multiscale approach ensures robust pedestrian detection across varying sizes and perspectives. The final feature representation \( F_{\text{final}} \) is obtained via fusion:  

\begin{equation}
F_{\text{final}} = \mathbf{w}^T \cdot \mathbf{F}
\end{equation}

where \( w_i \) are learnable weights, balancing contributions from different scales to enhance adaptability across detection scenarios.  

\subsection{Multi-Scale Spectral Feature Perception Module (MSFPM)}
The MSFPM plays a critical role in addressing the challenges of pedestrian detection under complex illumination conditions. Specifically designed to leverage the complementary information between RGB and infrared spectra, MSFPM effectively overcomes the limitations of existing methods in utilizing multimodal advantages in challenging scenarios.
The core of multimodal feature fusion is to effectively integrate RGB and infrared features while maximizing their complementary strengths across diverse environments. MSFPM achieves this by stacking features along the depth dimension and dynamically adjusting their contribution weights:

\begin{equation}
F_{\text{fused}} = w_{\text{RGB}} F_{\text{RGB}} + w_{\text{IR}} F_{\text{IR}}
\end{equation}

where \( w_{\text{RGB}} \) and \( w_{\text{IR}} \) are adaptive weights that prioritize RGB features in well-lit conditions and infrared features in low-light scenarios, ensuring robustness.

After fusion, MSFPM applies dimensionality reduction to enhance computational efficiency while preserving semantic richness:

\begin{equation}
F_{\text{reduced}} = W_{\text{reduce}} F_{\text{fused}}
\end{equation}

where \( W_{\text{reduce}} \) is a learnable projection matrix that compresses feature dimensions while maintaining cross-modal synergy. The SiLU activation function is then applied to improve non-linear representation and robustness:

\begin{equation}
F_{\text{activated}} = F_{\text{reduced}} \cdot \sigma(F_{\text{reduced}})
\end{equation}

where \( \sigma(x) = x \cdot \text{sigmoid}(x) \). Finally, the fused features undergo multi-scale convolutional processing for pedestrian detection, leveraging modality complementarity to ensure accurate results under varying illumination conditions.

\subsection{Illumination Robustness Feature Decoupling Module}
Feature decoupling \cite{34} aims to decompose latent representations into independent factors, reducing redundancy and enhancing feature clarity. To improve detection accuracy, we propose the Illumination Robustness Feature Decoupling Module (IRFDM), which separates visible light image features into human-related and background-related components. This minimizes background interference, refining pedestrian detection.  

As shown in the middle of Figure \ref{fig:main}, IRFDM employs two parallel processing paths to handle visible (\( f_v \)) and infrared (\( f_i \)) features separately, ensuring robust feature extraction under varying illumination conditions.

\paragraph{Process of Visible Light Images ($f_v$).}
The IRFDM process enhances feature separation by sequentially applying pooling, convolution, and activation functions. Max Pooling highlights key human-related features, while Average Pooling captures background details. Then, \( 1 \times 1 \) convolutions refine feature representations:

\begin{equation}
F' = W_{1\times1} * F
\end{equation}

where \( W_{1\times1} \) is the learnable weight matrix. Convolution and pooling layers further reduce feature dimensionality while improving stability.

Finally, a three-layer Multi-Layer Perceptron (MLP) refines the separation, producing decoupled human-related (\( f_v^{H} \)) and background-related (\( f_v^{B} \)) features:

\begin{equation}
f_v = f_v^{H} + f_v^{B}
\end{equation}

This structured decoupling minimizes background interference, enhancing pedestrian detection accuracy.

\paragraph{Process of Visible Light Images ($f_i$).}

The process begins with a Multi-Layer Perceptron (MLP), which captures complex nonlinear relationships for fine-grained feature differentiation. Max Pooling then reduces spatial dimensions while preserving the most relevant information. Next, the Sigmoid activation function normalizes feature values to \([0,1]\):
Finally, two \( 1 \times 1 \) convolutional layers separate features into human-related (\( f_v^{H} \)) and background-related (\( f_v^{B} \)) components:

\begin{equation}
f_v^{H}, f_v^{B} = W_H * F_{\text{norm}}, \quad W_B * F_{\text{norm}}
\end{equation}

where \( W_H \) and \( W_B \) are learnable transformation matrices. This structured feature decoupling enhances pedestrian detection by reducing background interference.

\paragraph{Orthogonal Regularization Loss Function.}
To ensure the independence of decoupled human and background features, an orthogonal regularization method be applied. This method involves adding a regularization term to the loss function to penalize the similarity between human features and background features. Specifically, this can be achieved by minimizing the dot product between these two sets of feature vectors, as orthogonal or uncorrelated vectors have a dot product of zero. The loss function can be formulated as:

\begin{equation}
\mathbf{L}_{\mathbf{con}} = \sum_{\mathbf{i} = 1}^{\mathbf{N}} \left| \mathbf{f}_{\mathbf{v}, \mathbf{i}}^{\mathbf{H}} \cdot \mathbf{f}_{\mathbf{v}, \mathbf{i}}^{\mathbf{B}} \right|^{2} + \mathbf{\lambda} \sum_{\mathbf{i} = 1}^{\mathbf{N}} \left| \mathbf{f}_{\mathbf{i}, \mathbf{i}}^{\mathbf{H}} \cdot \mathbf{f}_{\mathbf{i}, \mathbf{i}}^{\mathbf{B}} \right|^{2}
\end{equation}

where

\begin{itemize}
	\item $\mathbf{f}_{\mathbf{v,i}}^{\mathbf{H}}$ and $\mathbf{f}_{\mathbf{v,i}}^{\mathbf{B}}$ are the decoupled human and background features from visible light images, respectively.
	\item $\mathbf{f}_{\mathbf{i,i}}^{\mathbf{H}}$ and $\mathbf{f}_{\mathbf{i,i}}^{\mathbf{B}}$ are the decoupled human and background features from infrared images, respectively.
	\item $\mathbf{\lambda}$ is a hyperparameter that controls the weight of the infrared images term.
	\item $\mathbf{N}$ is the number of samples. Each sample $\mathbf{i}$ contributes to the overall loss, ensuring that the regularization is applied across all the data.
	\item The regularization terms $\mathbf{f}_{\mathbf{v,i}}^{\mathbf{H}} \cdot \mathbf{f}_{\mathbf{v,i}}^{\mathbf{B}}$ and $\mathbf{f}_{\mathbf{i,i}}^{\mathbf{H}} \cdot \mathbf{f}_{\mathbf{i,i}}^{\mathbf{B}}$ aim to minimize the correlation between human features and background features within the same image modality.
\end{itemize}

This approach not only effectively separates the features but also ensures their independence, which is crucial for improving the accuracy and robustness of object detection systems.

\subsection{Contrastive Learning Paradigm}
After decoupling human and background features, contrastive learning is essential to further enhance feature discrimination. This approach strengthens the model’s ability to distinguish humans from the background, improving recognition and classification across diverse lighting and environmental conditions.

\begin{figure}
    \centering
    \includegraphics[width=\linewidth]{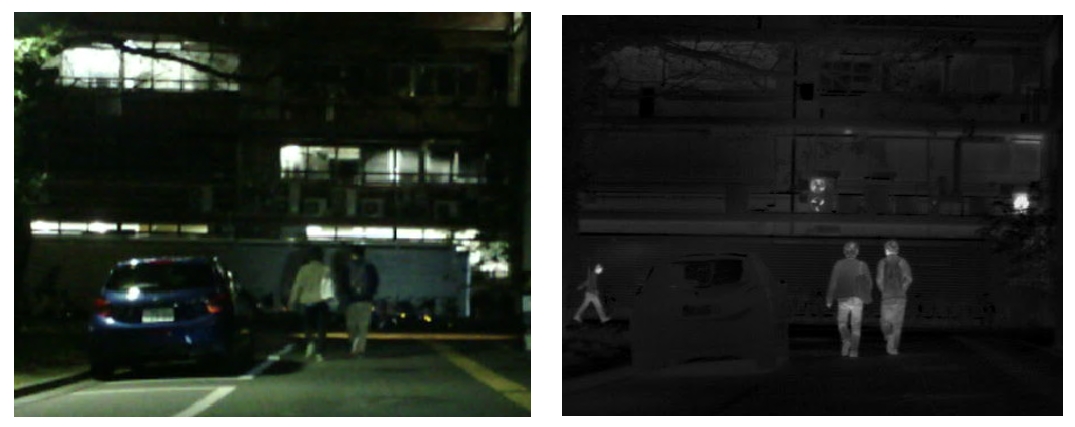}
    \caption{Example of RGB and Infrared Image Pairs. Left: RGB Image. Right: Infrared Image.}
    \label{fig:rgb_vs_ir}
\end{figure}

\paragraph{Selection of Visible Light and Infrared Image Pairs.}
We select visible and infrared image pairs from the MSDS and LLVIP Datasets (Figure \ref{fig:rgb_vs_ir}) under similar scene conditions, enabling the model to recognize pedestrians across lighting variations. Careful pair selection ensures high-quality positive and negative samples, crucial for effective contrastive learning.

For accurate multimodal pedestrian detection, visible and infrared image pairs must be strictly aligned in targets, background, and dimensions, sharing unified labels for seamless feature fusion. This alignment minimizes modality discrepancies and enhances robustness.  

\begin{figure}
	\centering
	\includegraphics[width=\linewidth]{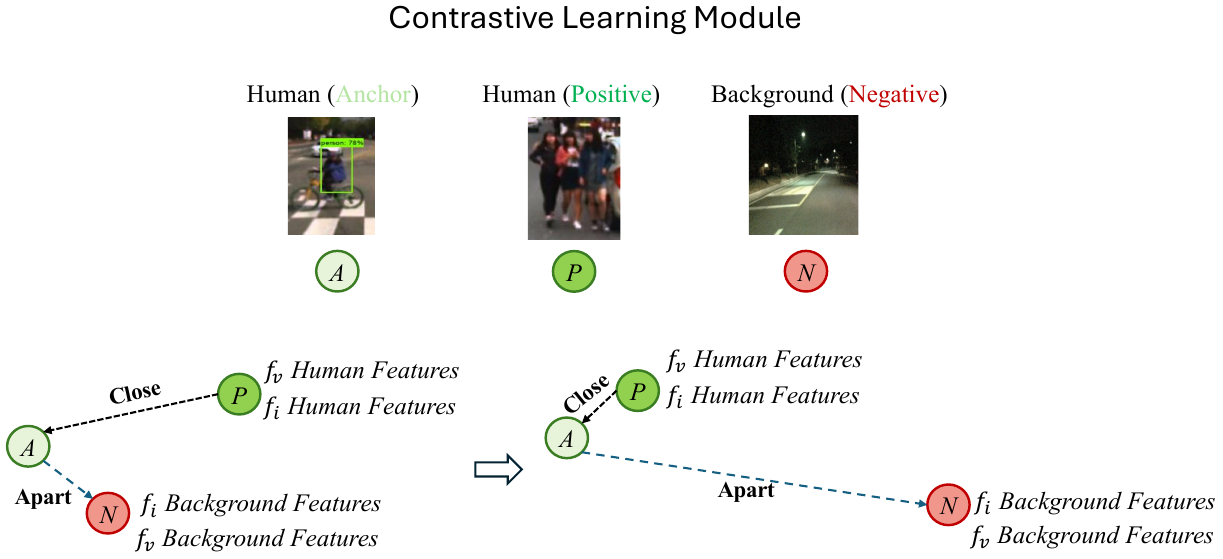}
	\caption{Contrastive learning paradigm.}
	\label{fig:cl}
\end{figure}

We implement a contrastive learning framework (Figure \ref{fig:cl}) using visible and infrared features to distinguish humans from the background. Triplet Loss optimizes feature learning by maximizing inter-class distances and minimizing intra-class distances, with a margin term defining decision boundaries. The model’s core components—Anchor (A), Positive (P), and Negative (N)—form the basis of the Triplet Loss function, ensuring effective feature discrimination.

\paragraph{Triplet Loss Function.}
The triplet loss is used in the contrastive learning framework to train the model by increasing the distance between dissimilar features (negative samples) and decreasing the distance between similar features (positive samples). The triplet loss function is defined as:
\[
\mathbf{L} = \sum_{\mathbf{i} = \mathbf{1}}^{\mathbf{N}} \max\left( \mathbf{d}\left( \mathbf{f}_{\mathbf{a,i}}, \mathbf{f}_{\mathbf{p,i}} \right) - \mathbf{d}\left( \mathbf{f}_{\mathbf{a,i}}, \mathbf{f}_{\mathbf{n,i}} \right) + \text{margin}, 0 \right)
\]

Where:

\begin{itemize}
	\item $\mathbf{f}_{\mathbf{a,i}}$ is the feature vector of the anchor sample for the i-th triplet.
	\item $\mathbf{f}_{\mathbf{p,i}}$ is the feature vector of the positive sample which is similar to the anchor.
	\item $\mathbf{f}_{\mathbf{n,i}}$ is the feature vector of the negative sample which is dissimilar to the anchor.
	\item $\mathbf{d}\left( \mathbf{x}, \mathbf{y} \right)$ represents the distance function, typically Euclidean distance, measuring the dissimilarity between two feature vectors.
	\item $\text{margin}$ is a predefined threshold that dictates the minimum difference required between the anchor-positive distance and the anchor-negative distance to encourage effective learning.
\end{itemize}

\paragraph{For Background Features.}
For background features, the positive sample is the background-related feature, while the negative sample is the human feature (background-unrelated). Through contrastive learning, we minimize the distance between all positive samples (visible background and infrared background) and maximize the distance between positive and negative samples. This approach enhances the model to learn background features more effectively, thereby improving the distinction between human and background features.

To minimize the distance between positive samples ($\mathbf{f}_{\mathbf{v}}$ background features to $\mathbf{f}_{\mathbf{i}}$ background features) and maximize the distance between negative samples and the anchor (Background to human), thereby enhancing the model's ability to discern background features difference apart from human features. The triplet loss for background features can be mathematically represented as:
\[
L_{\text{bg}} = \sum_{i = 1}^{N} \max\left( d\left( f_{a,i}^{B}, f_{p,i}^{B} \right) - d\left( f_{a,i}^{B}, f_{n,i}^{H} \right) + \text{margin}, 0 \right)
\]

\paragraph{For Human Features.}
Similarly, for human features, the positive sample is the human-related feature, and the negative sample is the background feature(human-unrelated). Through contrastive learning, we minimize the distance between all positive samples (human in visible images and human in infrared images), enabling the model to learn human features more effectively. By maximizing the distance between positive and negative samples, the model improves its ability to distinguish between human and background features.

To reduce the feature distance among human instances (positive samples) while expanding the gap between humans and their backgrounds (negative samples), thus facilitating a clearer distinction between human and environmental features. The triplet loss for human features can be mathematically represented as:
\[
L_{\text{human}} = \sum_{i = 1}^{N} \max\left( d\left( f_{a,i}^{H}, f_{p,i}^{H} \right) - d\left( f_{a,i}^{H}, f_{n,i}^{B} \right) + \text{margin}, 0 \right)
\]
\subsection{Detection Head}
The category prediction is responsible for classifying detected objects and outputting their category probabilities. As this study focuses on pedestrian detection, the module specifically predicts the probability of an object belonging to the pedestrian category. To achieve this, a fully connected layer is employed, followed by a softmax function to convert the raw predictions into a probability distribution. The formulation for category prediction is as follows:
\[
C_{\text{pred}} = \text{Softmax}\left( W_{c} \cdot F + b_{c} \right)
\]
Where $W_{c}$ is the weight matrix for category prediction, $F$ is the fused multimodal feature matrix, $b_{c}$ is the bias term, and $\text{Softmax}$ is the activation function used to calculate the probability of the target belonging to each category.

By integrating features from both RGB and infrared spectra, the category prediction module effectively addresses classification tasks under complex illumination conditions. This integration significantly enhances the accuracy of pedestrian detection by reducing false positives in low-light or cluttered background scenarios.

\paragraph{Bounding Box Prediction.}
The bounding box prediction is responsible for localizing detected objects by predicting their center coordinates $(x, y)$, width $w$, and height $h$. An anchor-based mechanism and regression to refine the bounding box parameters. The bounding box prediction is formulated as follows:

\[
B_{\text{pred}} = \left( \sigma\left( t_{x} \right) + x_{\text{cell}}, \sigma\left( t_{y} \right) + y_{\text{cell}}, p_{w}e^{t_{w}}, p_{h}e^{t_{h}} \right)
\]

Where $(t_{x}, t_{y}, t_{w}, t_{h})$ are the predicted adjustment parameters for the bounding box, $(x_{\text{cell}}, y_{\text{cell}})$ are the top-left coordinates of the grid cell, $(p_{w}, p_{h})$ are the width and height of the anchor box, and $\sigma(\cdot)$ is the Sigmoid activation function, which constrains the coordinates to the range $[0, 1]$.

\paragraph{Confidence Prediction.}
The confidence prediction evaluates whether a detected bounding box contains a pedestrian and assesses the precision of the box. The confidence score is determined by the probability of the object's presence, $P_{\text{obj}}$, and the Intersection over Union (IoU) between the predicted bounding box and the ground truth box. The formula is expressed as:

\[
S_{\text{conf}} = P_{\text{obj}} \times \text{IoU}\left( B_{\text{pred}}, B_{\text{true}} \right)
\]

Where $S_{\text{conf}}$ is the probability that the predicted bounding box contains a target object, and $\text{IoU}\left( B_{\text{pred}}, B_{\text{true}} \right)$ measures the overlap between the predicted bounding box $B_{\text{pred}}$ and the ground truth box $B_{\text{true}}$.

\begin{algorithm}[t] \caption{PedDet Algorithm Overview} \label{alg:overview} \begin{algorithmic}[1] \REQUIRE RGB image $I_{rgb}$, Infrared image $I_{ir}$ \ENSURE Detection results $B_{pred}$, $C_{pred}$, $S_{conf}$ \STATE \textbf{Feature Extraction:} \STATE $F_{rgb} \leftarrow \text{YOLOv10}(I_{rgb})$ \COMMENT{Generate multi-scale RGB features} \STATE $F_{ir} \leftarrow \text{YOLOv10}(I_{ir})$ \COMMENT{Generate multi-scale IR features} \STATE \textbf{MSFPM Processing:} \STATE $F_{fused} \leftarrow w_{rgb}F_{rgb} + w_{ir}F_{ir}$ \COMMENT{Dynamic fusion} \STATE $F_{reduced} \leftarrow W_{reduce}F_{fused}$ \COMMENT{Dimensionality reduction} \STATE $F_{activated} \leftarrow \text{SiLU}(F_{reduced})$ \STATE \textbf{IRFDM Processing:} \STATE $f_v^H, f_v^B \leftarrow \text{DecoupleFeatures}(F_{activated})$ \COMMENT{Visible branch} \STATE $f_i^H, f_i^B \leftarrow \text{DecoupleFeatures}(F_{activated})$ \COMMENT{Infrared branch} \STATE \textbf{Contrastive Learning:} \STATE $\mathcal{L}_{bg} \leftarrow \text{ComputeTripletLoss}(f_v^B, f_i^B, f_v^H)$ \STATE $\mathcal{L}_{human} \leftarrow \text{ComputeTripletLoss}(f_v^H, f_i^H, f_v^B)$ \STATE \textbf{Detection Head:} \STATE $C_{pred} \leftarrow \text{Softmax}(W_cF_{activated} + b_c)$ \STATE $B_{pred} \leftarrow \text{DecodeBox}(t_x, t_y, t_w, t_h)$ \STATE $S_{conf} \leftarrow P_{obj} \times \text{IoU}(B_{pred}, B_{true})$ \STATE \textbf{Joint Optimization:} \STATE $\mathcal{L}_{total} = \mathcal{L}_{cls} + \mathcal{L}_{box} + \mathcal{L}_{conf} + \lambda_1\mathcal{L}_{con} + \lambda_2(\mathcal{L}_{bg} + \mathcal{L}_{human})$ \end{algorithmic} \end{algorithm}

By leveraging the enhanced features from the IRFDM, the confidence prediction effectively differentiates targets from complex backgrounds. This enhanced capability ensures accurate confidence scores even under challenging conditions, such as low-light or high-brightness environments, maintaining robustness and reliability in pedestrian detection tasks.

\paragraph{Joint Loss Function.}
The overall prediction loss function integrates multiple components, including regularization constraints, the contrastive learning paradigm, classification, bounding box regression, and confidence prediction losses. To achieve comprehensive optimization of the model, we define a joint loss function, $L_{\text{total}}$, as follows:

\[
L_{\text{total}} = L_{\text{cls}} + L_{\text{box}} + L_{\text{conf}} + \lambda_{1}L_{\text{con}} + \lambda_{2}\left( L_{\text{bg}} + L_{\text{human}} \right)
\]

where $L_{\text{cls}}$ is the classification loss, $L_{\text{box}}$ is the localization loss, $L_{\text{conf}}$ is the confidence loss, $L_{\text{con}}$ is the decoupling constraint loss, and $L_{\text{bg}} + L_{\text{human}}$ is the contrastive loss. The terms $\lambda_{1}$ and $\lambda_{2}$ are hyperparameters that control the relative importance of the decoupling constraint loss and the contrastive loss, respectively.

The weights of these loss components can be adjusted to balance model performance. The choice of hyperparameters $\lambda_{1}$ and $\lambda_{2}$ depends on the specific characteristics of the dataset and the task requirements. By optimizing all these components, the overall prediction loss $L_{\text{total}}$ ensures that the model performs robustly across all detection tasks. This design enables accurate and reliable pedestrian detection, even in complex illumination conditions and diverse scenarios.

\section{Experiments}
\subsection{Experimental Setup and Datasets}
 \paragraph{Setup.}
 During the training phase, we adopted transfer learning by initializing the model with pre-trained weights from the MS COCO dataset \cite{lin2014microsoft}. By leveraging these pre-trained weights, the network could utilize general features from a large-scale dataset, significantly reducing training time and improving initial convergence speed. During validation, all input images were resized to a consistent resolution of 256$\times$256 pixels. This resizing ensures uniformity in the input data, minimizing computational overhead while maintaining detection accuracy. The validation process strictly adhered to the hyperparameter configurations used during training, ensuring the reliability and consistency of evaluation results. For training and validation configuration and hyperparameters, see \textbf{Supplementary Section 2: Implement Details}.

\paragraph{Datasets.}
We primarily evaluated our model on two benchmark datasets: the MSRS \cite{34} and LLVIP \cite{35} datasets. These datasets have become key benchmarks in the field as they reflect real-world scenarios. See \textbf{Supplementary Section 3: Datasets}.

\subsection{Comparative Study}

\begin{figure}[t]
	\centering
	\includegraphics[width=\linewidth]{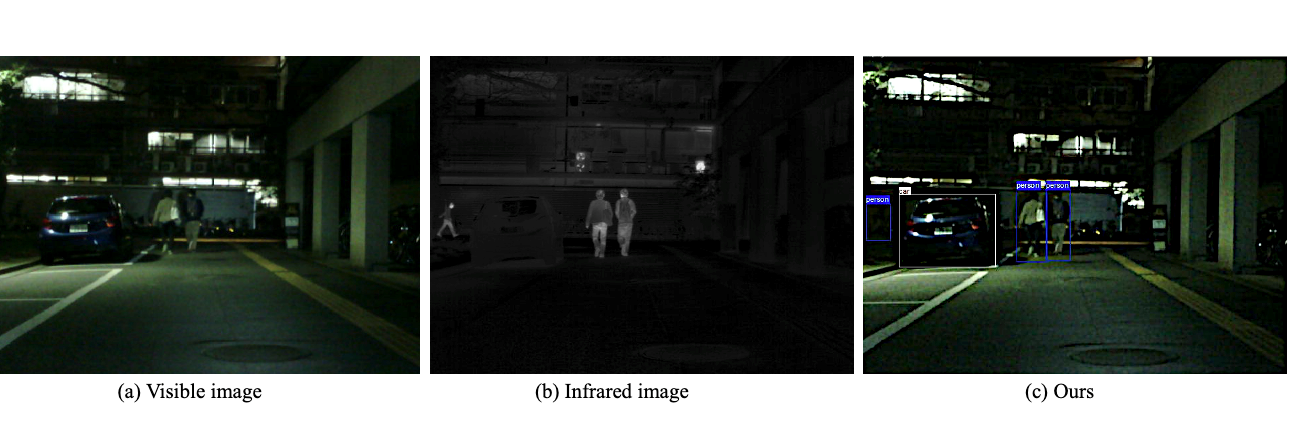}
	\caption{Comparison of traditional RGB-modal (a, b) and ours multi-modal (c) pedestrian detection result. (a) RGB image, (b) Infrared image, and (c) Detection result of the proposed multi-modal algorithm.}
	\label{fig:ours}
\end{figure}

To comprehensively evaluate the performance of PedDet, we compared its detection accuracy against several state-of-the-art multimodal pedestrian detection algorithms, including: SSD \cite{36}, RetinaNet \cite{1}, YOLOv10 \cite{4}, Faster R-CNN \cite{10}, DDQ-DETR \cite{37}, Halfway fusion \cite{9}, ProbEn \cite{18}, ARCNN-Extension \cite{38}, PoolFuser \cite{39}, LENFusion \cite{40}. The evaluation was conducted on two benchmark datasets, MSRS and LLVIP, with standard metrics including mean Average Precision (mAP) and its at 50\% Intersection over Union (mAP@50). The experimental results are shown in Table \ref{tab:comparison} below.

\begin{table}[H]
  \centering
\resizebox{\linewidth}{!}{
  \begin{tabular}{l c c c c c}
    \toprule
    Method & Backbone & Dataset & mAP@50$\uparrow$ & mAP$\uparrow$ & Modality \\
    \midrule

    \multicolumn{6}{c}{\textbf{Infrared}} \\
    \midrule
    SSD & VGG16 & LLVIP & 90.2 & 53.5 & Infrared \\
    RetinaNet & ResNet50 & LLVIP & 94.8 & 55.1 & Infrared \\
    YOLOv10 & CSPDarknet & LLVIP & 93.1 & 53.7 & Infrared \\
    Faster R-CNN & ResNet50 & LLVIP & 94.6 & 54.5 & Infrared \\
    DDQ-DETR & ResNet50 & LLVIP & 93.9 & 58.6 & Infrared \\
    \midrule

    \multicolumn{6}{c}{\textbf{RGB}} \\
    \midrule
    SSD & VGG16 & LLVIP & 82.6 & 39.8 & RGB \\
    RetinaNet & ResNet50 & LLVIP & 88.0 & 42.8 & RGB \\
    YOLOv10 & CSPDarknet & LLVIP & 87.8 & 50.2 & RGB \\
    Faster R-CNN & ResNet50 & LLVIP & 87.0 & 47.5 & RGB \\
    DDQ-DETR & ResNet50 & LLVIP & 86.1 & 46.7 & RGB \\
    \midrule

    \multicolumn{6}{c}{\textbf{Infrared + RGB}} \\
    \midrule
    Halfway Fusion & VGG16 & LLVIP & 91.4 & 55.1 & Infrared + RGB \\
    ProbEn & ResNet50 & LLVIP & 93.4 & 51.5 & Infrared + RGB \\
    ARCNN-Extension & VGG16 & LLVIP & 89.2 & 56.2 & Infrared + RGB \\
    PoolFuser & ResNet-34s & LLVIP & 80.3 & 38.4 & Infrared + RGB \\
    LENFusion & CSPDarknet & LLVIP & 81.6 & 53.0 & Infrared + RGB \\
    \rowcolor{yellow} \textbf{PedDet (Ours)} & CSPDarknet & LLVIP & \textbf{95.8} & \textbf{56.8} & Infrared + RGB \\
    \bottomrule
  \end{tabular}
}
\caption{\textbf{Comparison of detection performance of SOTA methods.} The best values are in \textbf{bold}.}
\label{tab:comparison}
\end{table}

\begin{figure}[h]
    \centering
    \begin{minipage}{0.48\linewidth}
    \centering
    \includegraphics[width=\linewidth]{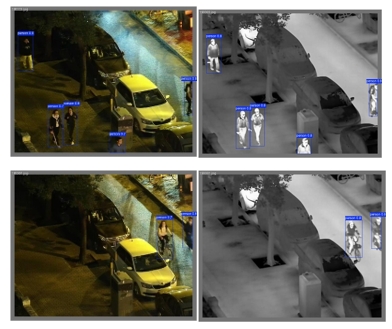}
    \end{minipage}
    \begin{minipage}{0.51\linewidth}
        \centering
    \includegraphics[width=\linewidth]{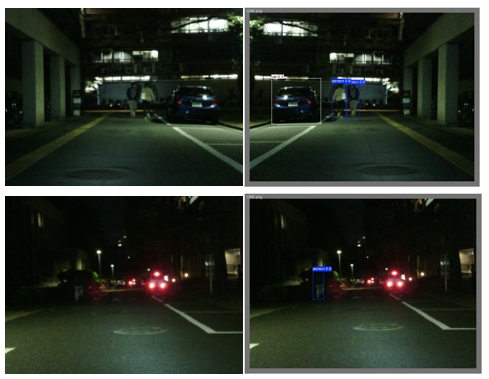}
    \end{minipage}
    \caption{Comparison of detection visualization effects with and without infrared assistance.}
    \label{fig:vis1}
\end{figure}

The detection results as shown in Figure \ref{fig:ours}(a), visible-light images exhibit challenges in distinguishing humans from the background under poor illumination conditions. This limitation is evident in the low contrast between objects and their surroundings, which makes pedestrian detection difficult. In contrast, infrared images provide distinct object contours, enabling easier differentiation of pedestrians from the background. The proposed PedDet effectively integrates the complementary features of visible and infrared spectra, highlighting pedestrians in the fused feature representation. This integration results in robust detection performance, particularly in low-light scenarios as shown in  Figure \ref{fig:vis1}.

\subsection{Ablation Study}
\paragraph{Effect of Contrastive Learning.}
In this section, we evaluate the impact of contrastive learning (CL) on the performance of different backbones. The results are presented in Table~\ref{tab:contrastive_learning}. As shown, the introduction of contrastive learning consistently improves the performance across all backbones. For instance, YOLOv10-S achieves a performance of 54.8  \% with CL, which is an improvement of 1.6  \% compared to its performance without CL. Similar improvements are observed for YOLOv10-M and YOLOv10-L, with performance gains of 1.6  \% and 1.7  \%, respectively. These results highlight the effectiveness of contrastive learning in enhancing the model's capability.

\begin{table}[h!]
\centering
\caption{Impact of different backbones on the task with and without contrastive learning.}
\label{tab:contrastive_learning}
\resizebox{0.9\linewidth}{!}{ %
\begin{tabular}{>{\centering\arraybackslash}p{2.5cm} |>{\centering\arraybackslash}p{2cm} >{\centering\arraybackslash}p{2.8cm}} 
\toprule
\textbf{Backbone} & \textbf{w/o CL} & \textbf{w/ CL} \\
\midrule
YOLOv10-S & 53.2  \% & 54.8  \%  \textcolor{red}{ ($\uparrow$ 1.6  \%)} \\
YOLOv10-M & 55.1  \% & 56.7  \%  \textcolor{red}{   ($\uparrow$ 1.6  \%)} \\
YOLOv10-L & 56.4  \% & 58.1  \%  \textcolor{red}{   ($\uparrow$ 1.7  \%)} \\
\bottomrule
\end{tabular}
}
\end{table}

\paragraph{Effect of Additional Modality.}
We further investigated the impact of different modalities (AM) on task performance, with the results summarized in Table~\ref{label3}. The inclusion of additional modalities significantly improved performance. After introducing MSFPM and IRFDM, PedDet achieved a performance of 95.8  \%, representing an improvement of 1.1  \% compared to when CL was not used. This demonstrates the importance of leveraging additional modalities to enhance the model's effectiveness.

\begin{table}[h!]
\centering
\caption{Performance comparison of different models.}
\label{label3}
\resizebox{0.9\linewidth}{!}{ %
\begin{tabular}{c|c c}
\toprule
\textbf{Models} & \textbf{mAP@50} & \textbf{mAP} \\
\midrule
\rowcolor{yellow} PedDet    & 95.8   & 56.8    \\
PedDet (w/o MSFPM)   & 93.4      & 54.3      \\
PedDet (w/o IRFDM)   & 94.1       & 55.0     \\
PedDet (w/o CL)   & 94.7      & 55.5      \\
PedDet (w/o MSFPM and IRFDM)   & 91.2      & 51.8      \\
PedDet (w/o MSFPM and CL)   & 92.8      & 53.1      \\
PedDet (w/o IRFDM and CL)   & 93.0      & 53.5       \\
\bottomrule
\end{tabular}
}
\end{table}

\paragraph{T-SNE Visualization of Feature Decoupling Results.}
To further demonstrate the effectiveness of our feature decoupling module, we performed t-SNE visualization. As shown in the Figure \ref{fig:decouple} below, the t-SNE visualization shows the model's ability to distinguish between \textit{human relevant (green color) and human irrelevant (red color)} features over different training epochs (100, 200, 500, 1000).  In the initial stages of training, the human-relevant feature and human-irrelevant feature points are highly mixed, indicating the model's initial weak capability in feature distinction.

However, as training progresses to 200 epochs, the data points begin to show some clustering trends, although there is still a lot of overlap. By 500 epochs, the separation between human-relevant features and human-irrelevant features becomes more clearer, indicating a significant improvement in the model's ability to discriminate features. At 1000 epochs, the clustering of human-relevant and human irrelevant features is very obvious, with minimal overlap, which shows that the model has effectively learned to differentiate between the two types of features, accurately separating and grouping them. 

This gradual improvement of the learning process indicates that the training strategy of the model is successful, enabling the model to accurately perform tasks such as classification, recognition in complex datasets.
\begin{figure}[h]
	\centering
	\includegraphics[width=\linewidth]{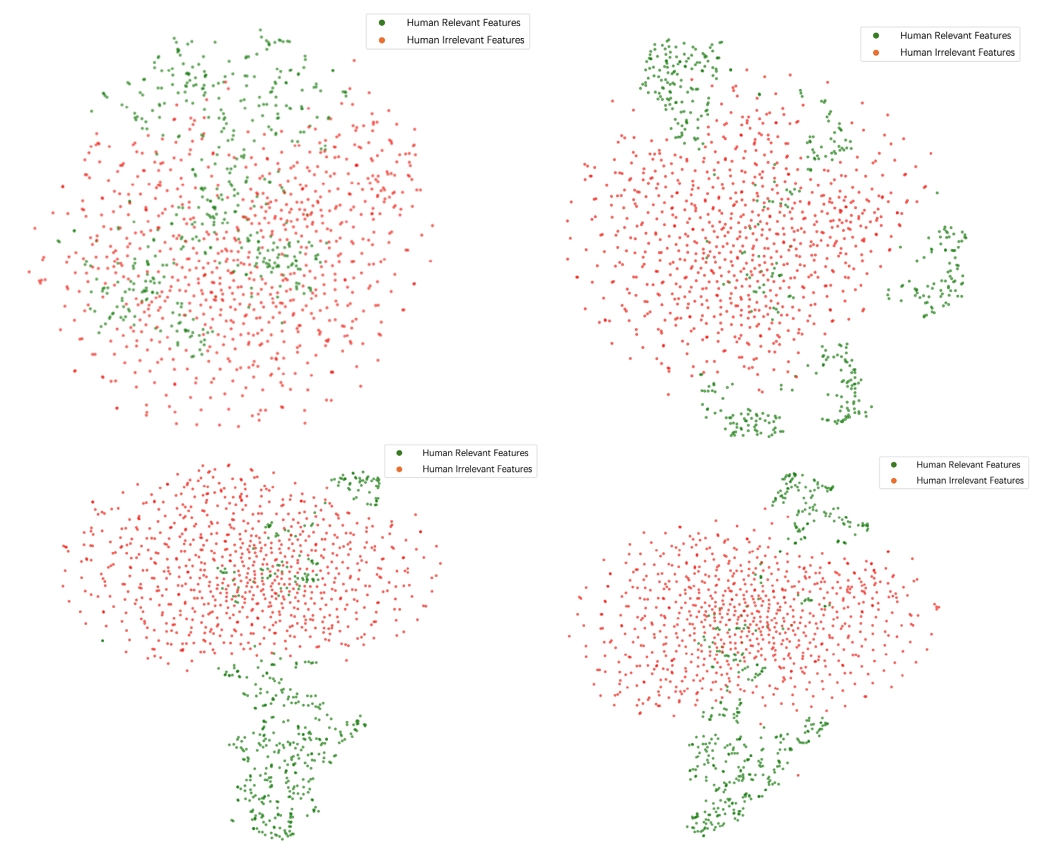}
	\caption{t-SNE visualization of feature decoupling results.}
	\label{fig:decouple}
\end{figure}

\section{Conclusions}

In this paper, we propose PedDet, an adaptive framework that enhances pedestrian detection under complex illumination by leveraging spectral complementarity. It integrates the MSFPM and IRFDM to optimize detection across visible and infrared spectra. MSFPM adjusts feature weights, while IRFDM mitigates illumination noise by decoupling pedestrian and background features. A contrastive learning approach further improves feature discrimination. Experimental results show PedDet outperforms existing methods, establishing it as a reliable and effective solution for multimodal pedestrian detection and a benchmark in intelligent transportation systems.

\clearpage

\newcommand{\supplementarytitle}{
    \twocolumn[
        \begin{center}
            \LARGE \textbf{PedDet: Adaptive Spectral Optimization for Multimodal Pedestrian Detection} \\
            \vspace{1em}
            \Large \textbf{- Supplementary Materials -}
            \vspace{1em}
        \end{center}
    ]
}

\supplementarytitle

\renewcommand{\thesection}{\arabic{section}}
\setcounter{section}{0}

\renewcommand{\thefigure}{\arabic{figure}}
\setcounter{figure}{0}

\renewcommand{\thetable}{\arabic{table}}
\setcounter{table}{0}

\section{Related Works}

\paragraph{Multimodal Fusion Challenges}
Multimodal information, such as RGB and infrared spectra, has been extensively studied for improving pedestrian detection performance. Despite its potential, many existing methods struggle to effectively harness the complementary advantages of multimodal features. The YOLO series, first introduced by Joseph et al. \cite{5}, has demonstrated excellent performance in single-modal detection tasks. However, its original architecture lacks specific mechanisms for multimodal fusion. Subsequent versions, such as YOLOv3 \cite{5} and its successors \cite{6}, focus primarily on structural optimization and speed improvements without addressing the challenges of multimodal integration.

To address these limitations, researchers have proposed various solutions. For example, Hwang et al. \cite{7} introduced the KAIST dataset, a large-scale multispectral pedestrian benchmark with well-aligned visible and thermal images, along with dense pedestrian annotations. They also proposed a method that extracts Aggregated Channel Features (ACF+T+THOG) and employs Boosted Decision Trees (BDT) for classification. However, this approach relies on shallow feature fusion, limiting its adaptability to environmental changes and resulting in instability under complex scenarios.

Wagner et al. \cite{8} pioneered the application of deep neural networks (DNNs) in multispectral pedestrian detection and compared early and late fusion strategies. Liu et al. \cite{9} extended this research by applying Faster R-CNN  \cite{10} to multispectral pedestrian detection and designing four ConvNet-based fusion architectures. Among these, the Halfway Fusion model, which merges middle-layer convolutional features from dual-branch ConvNets, achieved the best results. König et al. \cite{11} introduced the Fusion RPN+BDT model, integrating dual-stream DNNs at middle convolutional layers for enhanced fusion.

Recently, Park et al.  \cite{12} proposed a three-branch DNN architecture to handle multimodal inputs, incorporating a Channel Weighted Fusion (CWF) layer to enhance detection performance by adaptively weighting the contributions of each modality. Loveday et al.  \cite{13} developed an orthogonal dual-camera imaging system to capture parallax-free, well-aligned multispectral images. Their findings demonstrated that combining visible and infrared data significantly improves foreground object detection compared to using single-modal data.
\paragraph{Illumination Robustness Challenges}
With the advancement of object detection \cite{zhang2024meddet,cai2024medical,cai2024msdet}, pedestrian detection methods have divided into two main methods: one-stage detectors [14], and two-stage detectors [15-16],. However, these detectors predominantly rely on RGB images, leading to performance degradation under low-light or extreme illumination conditions.

The YOLO series \cite{6}, while excelling under standard lighting conditions, exhibits significant limitations in handling complex illumination scenarios. Models such as YOLOv5 and YOLOv6 improved detection speed and accuracy through structural optimizations but struggled to maintain robustness in extreme lighting environments. These algorithms often fail to effectively distinguish pedestrian features from background noise under conditions like uneven lighting or occlusion. Although RGB images provide rich texture and detail information, achieving reliable detection results in challenging conditions—such as extreme lighting, occlusions, and complex backgrounds—requires incorporating multimodal data.

Existing approaches have explored various fusion strategies for integrating multimodal information, including early fusion, late fusion, and intermediate fusion. Early fusion (pixel-level fusion) directly concatenates data from different modalities and processes it using conventional object detectors \cite{5,15,17}. Late fusion involves feeding each modality into separate single-modal detectors and subsequently merging the predicted bounding boxes using statistical methods \cite{18,19,20}. While simple, both early and late fusion methods often overlook the interdependencies between modalities, limiting their ability to fully exploit complementary features.

To address these limitations, recent studies have introduced \textit{illumination-aware feature fusion} and \textit{attention-based feature fusion} strategies. Illumination-aware fusion methods \cite{17,21,22,23} typically incorporate a classification branch to determine the significance of RGB features based on lighting conditions. However, classification-based approaches cannot accurately reflect the importance of individual regions within an image. In contrast, attention-based methods leverage spatial attention, channel attention, or cross-attention mechanisms derived from transformers to facilitate feature fusion \cite{9,24,25,26,27,28}. Spatial and channel attention generate element-wise and channel-wise weighting factors for multispectral features, respectively, while cross-attention models global contextual correlations to resolve feature misalignment between modalities. Although cross-attention provides superior fusion capabilities, it incurs high computational costs, posing challenges for real-time applications.

\section{Implement Details}

\paragraph{Epochs and Batch Size.} The training process was conducted for 50 epochs, with a batch size of 8 images per iteration.

\paragraph{Optimizer and Learning Rate.} We employed the Adam optimizer with an initial learning rate of 0.001. To address the issue of high learning rates during later training stages, we applied the Cosine Annealing learning rate scheduling strategy. This approach gradually decreases the learning rate throughout the training process, approaching zero at the end of each cycle. This dynamic decay effectively mitigates overfitting.

\paragraph{Weight Decay.} To further enhance model stability and generalization, a weight decay mechanism with a coefficient of 0.0005 was introduced. This regularization technique suppresses overly complex parameter updates in the network, improving the model’s ability to generalize to unseen data.
	
\paragraph{Warm-up Strategy.} To ensure a smooth transition into optimal training conditions, a warm-up strategy was applied during the first three epochs. This gradually increased the learning rate, preventing instability caused by large parameter adjustments in the early stages of training.
	
\paragraph{Multi-scale Training.} To improve the model's robustness and adaptability to varying object sizes, we incorporated a multi-scale training strategy. This involved randomly scaling images in different training batches, enhancing the network's ability to detect objects of various scales. This strategy is particularly advantageous for multimodal pedestrian detection, improving the detection of small objects while maintaining performance on larger targets.

The detailed configuration and hyperparameters are in Table \ref{tab:config}.

\begin{table}[h]
  \centering
\resizebox{\linewidth}{!}{
  \begin{tabular}{l c c}
    \toprule
    & \multicolumn{2}{c}{\textbf{Object Detectors}} \\
    \cmidrule(r){2-3}
    \textbf{Hyperparameter} & \textbf{Traditional YOLO} & \textbf{PedDet (Ours)} \\
    \midrule
    Learning Rate & 0.001 & 0.001 \\
    Epochs & 400 & 50 \\
    Batch Size & 16 & 8 \\
    Optimizer & SGD & AdamW \\
    Weight Decay & 0.0005 & 0.0005 \\
    Warmup Epochs & 3 & 5 \\
    Scheduler & Cosine & Cosine \\
    \bottomrule
  \end{tabular}
}
\caption{\textbf{Pedestrian detection training hyperparameters.} Comparison of hyperparameters used in traditional YOLO variants and our approach.}
\label{tab:config}
\end{table}

\begin{figure}[h]
    \centering
    \includegraphics[width=\linewidth]{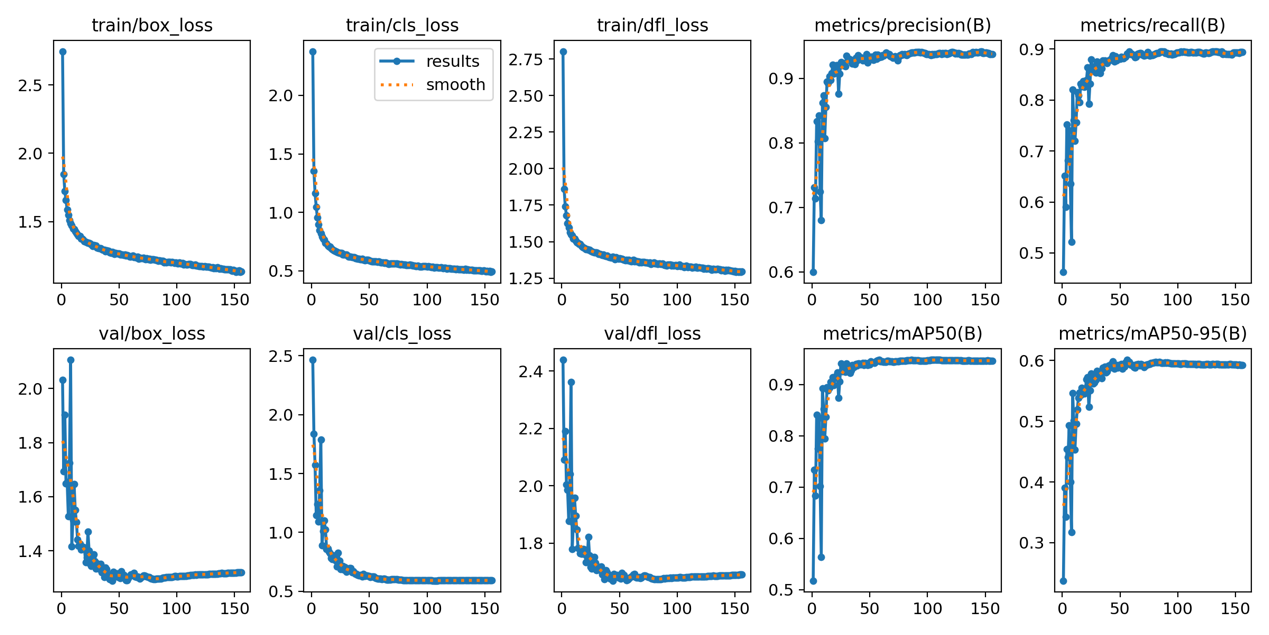}
    \caption{Training curves of LLVIP dataset.}
    \label{fig:fig_llvip}
    \includegraphics[width=\linewidth]{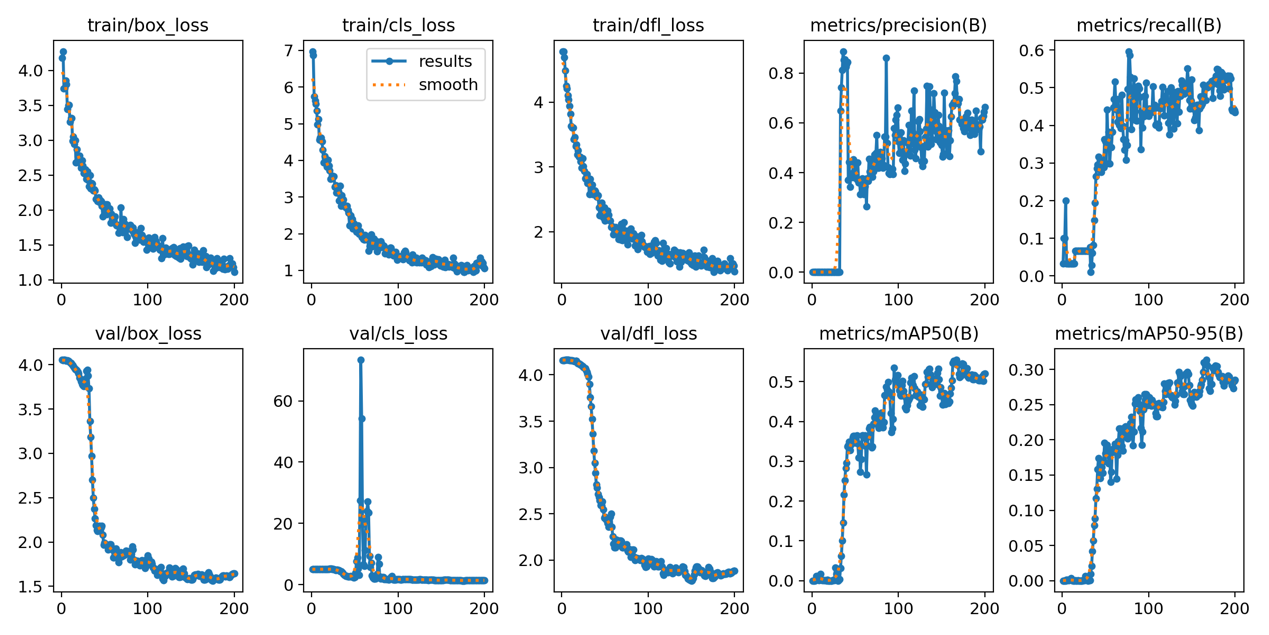}
    \caption{Training curves of MSRS dataset.}
    \label{fig:fig_msrs}
\end{figure}

\paragraph{Training Analysis.}
As shown in Figure \ref{fig:fig_llvip} and \ref{fig:fig_msrs}, in the evaluation of performance metrics, both precision and recall show a relatively fluctuating upward trend during the training process. As training progresses, both precision and recall gradually improve, indicating that the model becomes increasingly proficient at distinguishing targets from non-targets. The initial fluctuations observed in these metrics during certain training phases suggest the model's iterative adjustments to optimize its detection performance.

\section{Datasets}

\paragraph{Multi-Spectral Road Scenarios (MSRS).}
The MSRS dataset is a refined version of the MFNet dataset, specifically designed for infrared-visible image fusion in multimodal applications. The original MFNet dataset comprises 1,569 image pairs (820 captured during the day and 749 at night) with a resolution of 480$\times$640. However, it suffers from several limitations, including misaligned image pairs and low signal-to-noise ratio (SNR) and contrast in most infrared images.

To address these issues, the MSRS dataset enhances data quality through several steps:
\begin{itemize}
	\item \textbf{Removal of Misaligned Image Pairs:} 125 misaligned pairs were excluded, resulting in 715 daytime and 729 nighttime image pairs.
	\item \textbf{Infrared Image Enhancement:} A dark channel prior-based image enhancement algorithm was applied to improve the contrast and SNR of the infrared images.
\end{itemize}

As a result, the MSRS dataset comprises 1,444 pairs of high-quality, aligned infrared and visible images, making it a reliable benchmark for multimodal pedestrian detection under diverse lighting conditions.

\paragraph{LLVIP.}
The LLVIP dataset is a more challenging multimodal dataset, tailored for pedestrian detection under low-light conditions. It includes a large number of pedestrians captured in various lighting environments, making it particularly valuable for testing low-light detection models.

The dataset was collected with a dual-mode camera system that integrates visible and infrared cameras, ensuring temporal and spatial consistency between image pairs. Each pair was carefully registered and cropped to ensure identical fields of view and dimensions. This strict alignment makes the dataset especially suitable for image fusion and image-to-image translation tasks.

The LLVIP dataset comprises 12,025 pairs of aligned visible-infrared images, with a training set of 3,463 pairs. Each image has a resolution of 1,024$\times$1,280, providing high-quality data for training and evaluation.

\section{Visualization of Results}
\begin{figure}[h]
    \centering
    \includegraphics[width=\linewidth]{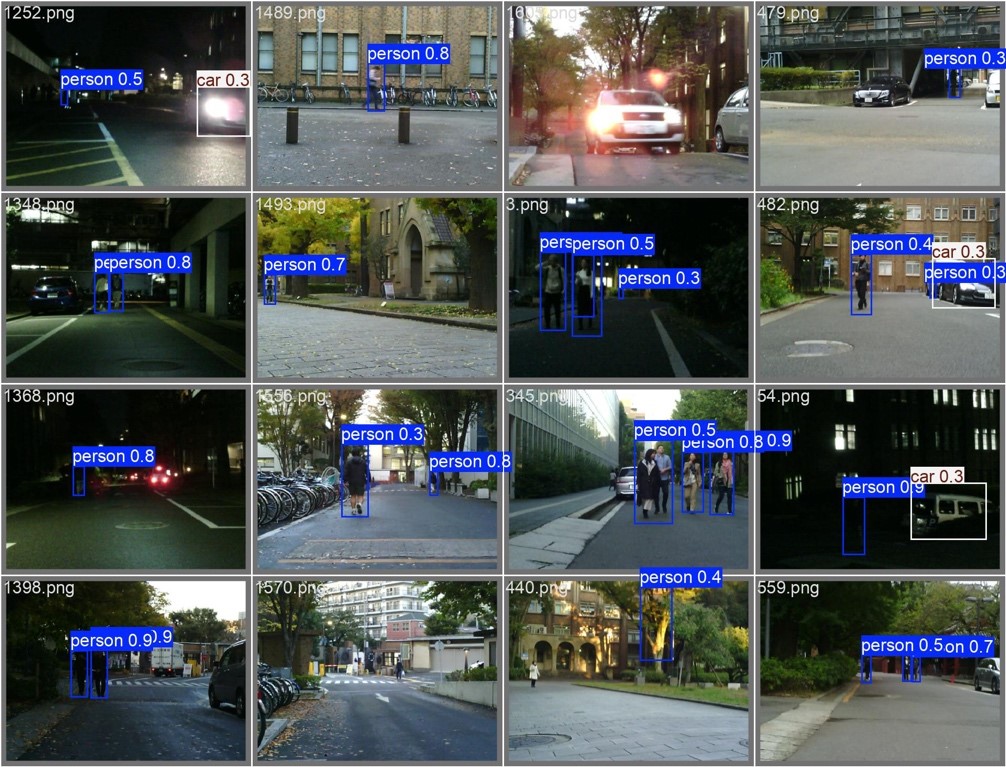}
    \caption{Visualization of PedDet results.}
    \label{fig:vis2}
\end{figure}

Figure \ref{fig:vis2} showcases typical examples of pedestrian detection in various environments, with each image highlighting the detected objects (pedestrians or vehicles) through bounding boxes and displaying the corresponding confidence scores. The detection results reflect the model's performance in handling complex visual scenes, especially its accuracy under different lighting conditions. In low-light or high-contrast environments, the model can still effectively identify pedestrians and exhibits strong robustness against background interference. This capability is made possible by the two core modules introduced in the paper: the Multi-scale Spectral Feature Perception Module (MSFPM) and the Illumination Robustness Feature Decoupling Module (IRFDM).

\clearpage
\begin{thebibliography}{}

\bibitem[\protect\citeauthoryear{Cai \bgroup \em et al.\egroup }{2024a}]{cai2024medical}
Guohui Cai, Ying Cai, Zeyu Zhang, Yuanzhouhan Cao, Lin Wu, Daji Ergu, Zhinbin Liao, and Yang Zhao.
\newblock Medical ai for early detection of lung cancer: A survey.
\newblock {\em arXiv preprint arXiv:2410.14769}, 2024.

\bibitem[\protect\citeauthoryear{Cai \bgroup \em et al.\egroup }{2024b}]{cai2024msdet}
Guohui Cai, Ying Cai, Zeyu Zhang, Daji Ergu, Yuanzhouhan Cao, Binbin Hu, Zhibin Liao, and Yang Zhao.
\newblock Msdet: Receptive field enhanced multiscale detection for tiny pulmonary nodule.
\newblock {\em arXiv preprint arXiv:2409.14028}, 2024.

\bibitem[\protect\citeauthoryear{Cao \bgroup \em et al.\egroup }{2023}]{39}
Yue Cao, Yanshuo Fan, Junchi Bin, and Zheng Liu.
\newblock Lightweight transformer for multi-modal object detection (student abstract).
\newblock In {\em Proceedings of the AAAI Conference on Artificial Intelligence}, volume~37, pages 16172--16173, 2023.

\bibitem[\protect\citeauthoryear{Chen \bgroup \em et al.\egroup }{2022}]{18}
Yi-Ting Chen, Jinghao Shi, Zelin Ye, Christoph Mertz, Deva Ramanan, and Shu Kong.
\newblock Multimodal object detection via probabilistic ensembling.
\newblock In {\em European Conference on Computer Vision}, pages 139--158. Springer, 2022.

\bibitem[\protect\citeauthoryear{Chen \bgroup \em et al.\egroup }{2024}]{40}
Jun Chen, Liling Yang, Wei Liu, Xin Tian, and Jiayi Ma.
\newblock Lenfusion: A joint low-light enhancement and fusion network for nighttime infrared and visible image fusion.
\newblock {\em IEEE Transactions on Instrumentation and Measurement}, 2024.

\bibitem[\protect\citeauthoryear{Hussain}{2023}]{6}
Muhammad Hussain.
\newblock Yolo-v1 to yolo-v8, the rise of yolo and its complementary nature toward digital manufacturing and industrial defect detection.
\newblock {\em Machines}, 11(7):677, 2023.

\bibitem[\protect\citeauthoryear{Hwang \bgroup \em et al.\egroup }{2015}]{7}
Soonmin Hwang, Jaesik Park, Namil Kim, Yukyung Choi, and In~So~Kweon.
\newblock Multispectral pedestrian detection: Benchmark dataset and baseline.
\newblock In {\em Proceedings of the IEEE conference on computer vision and pattern recognition}, pages 1037--1045, 2015.

\bibitem[\protect\citeauthoryear{Jia \bgroup \em et al.\egroup }{2021}]{35}
Xinyu Jia, Chuang Zhu, Minzhen Li, Wenqi Tang, and Wenli Zhou.
\newblock Llvip: A visible-infrared paired dataset for low-light vision.
\newblock In {\em Proceedings of the IEEE/CVF international conference on computer vision}, pages 3496--3504, 2021.

\bibitem[\protect\citeauthoryear{{Jocher} \bgroup \em et al.\egroup }{2022}]{2}
Glenn {Jocher}, Ayush {Chaurasia}, Alex {Stoken}, Jirka {Borovec}, {NanoCode012}, Yonghye {Kwon}, {TaoXie}, Kalen {Michael}, Jiacong {Fang}, {Imyhxy}, {Lorna}, Colin {Wong}, ''Zeng~Yifu'' {Zeng Yifu}, Abhiram {V}, Diego {Montes}, Zhiqiang {Wang}, Cristi {Fati}, Jebastin {Nadar}, {Laughing}, {UnglvKitDe}, {Tkianai}, {YxNONG}, Piotr {Skalski}, Adam {Hogan}, Max {Strobel}, Mrinal {Jain}, Lorenzo {Mammana}, and {Xylieong}.
\newblock {ultralytics/yolov5: v6.2 - YOLOv5 Classification Models, Apple M1, Reproducibility, ClearML and Deci.ai integrations}, August 2022.

\bibitem[\protect\citeauthoryear{Konig \bgroup \em et al.\egroup }{2017}]{11}
Daniel Konig, Michael Adam, Christian Jarvers, Georg Layher, Heiko Neumann, and Michael Teutsch.
\newblock Fully convolutional region proposal networks for multispectral person detection.
\newblock In {\em Proceedings of the IEEE conference on computer vision and pattern recognition workshops}, pages 49--56, 2017.

\bibitem[\protect\citeauthoryear{Li \bgroup \em et al.\egroup }{2018}]{19}
Chengyang Li, Dan Song, Ruofeng Tong, and Min Tang.
\newblock Multispectral pedestrian detection via simultaneous detection and segmentation.
\newblock {\em arXiv preprint arXiv:1808.04818}, 2018.

\bibitem[\protect\citeauthoryear{Li \bgroup \em et al.\egroup }{2020}]{21}
Xiang Li, Wenhai Wang, Lijun Wu, Shuo Chen, Xiaolin Hu, Jun Li, Jinhui Tang, and Jian Yang.
\newblock Generalized focal loss: Learning qualified and distributed bounding boxes for dense object detection.
\newblock {\em Advances in Neural Information Processing Systems}, 33:21002--21012, 2020.

\bibitem[\protect\citeauthoryear{Liang \bgroup \em et al.\egroup }{2022}]{22}
Tingting Liang, Xiaojie Chu, Yudong Liu, Yongtao Wang, Zhi Tang, Wei Chu, Jingdong Chen, and Haibin Ling.
\newblock Cbnet: A composite backbone network architecture for object detection.
\newblock {\em IEEE Transactions on Image Processing}, 31:6893--6906, 2022.

\bibitem[\protect\citeauthoryear{Lin \bgroup \em et al.\egroup }{2014a}]{lin2014microsoft}
Tsung-Yi Lin, Michael Maire, Serge Belongie, James Hays, Pietro Perona, Deva Ramanan, Piotr Doll{\'a}r, and C~Lawrence Zitnick.
\newblock Microsoft coco: Common objects in context.
\newblock In {\em Computer Vision--ECCV 2014: 13th European Conference, Zurich, Switzerland, September 6-12, 2014, Proceedings, Part V 13}, pages 740--755. Springer, 2014.

\bibitem[\protect\citeauthoryear{Lin \bgroup \em et al.\egroup }{2014b}]{24}
Tsung-Yi Lin, Michael Maire, Serge Belongie, James Hays, Pietro Perona, Deva Ramanan, Piotr Doll{\'a}r, and C~Lawrence Zitnick.
\newblock Microsoft coco: Common objects in context.
\newblock In {\em Computer Vision--ECCV 2014: 13th European Conference, Zurich, Switzerland, September 6-12, 2014, Proceedings, Part V 13}, pages 740--755. Springer, 2014.

\bibitem[\protect\citeauthoryear{Lin \bgroup \em et al.\egroup }{2017}]{17}
Tsung-Yi Lin, Piotr Doll{\'a}r, Ross Girshick, Kaiming He, Bharath Hariharan, and Serge Belongie.
\newblock Feature pyramid networks for object detection.
\newblock In {\em Proceedings of the IEEE conference on computer vision and pattern recognition}, pages 2117--2125, 2017.

\bibitem[\protect\citeauthoryear{Lin \bgroup \em et al.\egroup }{2020}]{23}
Matthieu Lin, Chuming Li, Xingyuan Bu, Ming Sun, Chen Lin, Junjie Yan, Wanli Ouyang, and Zhidong Deng.
\newblock Detr for crowd pedestrian detection.
\newblock {\em arXiv preprint arXiv:2012.06785}, 2020.

\bibitem[\protect\citeauthoryear{Liu \bgroup \em et al.\egroup }{2016a}]{9}
Jingjing Liu, Shaoting Zhang, Shu Wang, and Dimitris~N Metaxas.
\newblock Multispectral deep neural networks for pedestrian detection.
\newblock {\em arXiv preprint arXiv:1611.02644}, 2016.

\bibitem[\protect\citeauthoryear{Liu \bgroup \em et al.\egroup }{2016b}]{36}
Wei Liu, Dragomir Anguelov, Dumitru Erhan, Christian Szegedy, Scott Reed, Cheng-Yang Fu, and Alexander~C Berg.
\newblock Ssd: Single shot multibox detector.
\newblock In {\em Computer Vision--ECCV 2016: 14th European Conference, Amsterdam, The Netherlands, October 11--14, 2016, Proceedings, Part I 14}, pages 21--37. Springer, 2016.

\bibitem[\protect\citeauthoryear{Liu \bgroup \em et al.\egroup }{2021}]{26}
Ze~Liu, Yutong Lin, Yue Cao, Han Hu, Yixuan Wei, Zheng Zhang, Stephen Lin, and Baining Guo.
\newblock Swin transformer: Hierarchical vision transformer using shifted windows.
\newblock In {\em Proceedings of the IEEE/CVF international conference on computer vision}, pages 10012--10022, 2021.

\bibitem[\protect\citeauthoryear{Liu \bgroup \em et al.\egroup }{2022}]{25}
Jinyuan Liu, Xin Fan, Zhanbo Huang, Guanyao Wu, Risheng Liu, Wei Zhong, and Zhongxuan Luo.
\newblock Target-aware dual adversarial learning and a multi-scenario multi-modality benchmark to fuse infrared and visible for object detection.
\newblock In {\em Proceedings of the IEEE/CVF conference on computer vision and pattern recognition}, pages 5802--5811, 2022.

\bibitem[\protect\citeauthoryear{Loveday and Breckon}{2018}]{13}
Michael Loveday and Toby~P Breckon.
\newblock On the impact of parallax free colour and infrared image co-registration to fused illumination invariant adaptive background modelling.
\newblock In {\em Proceedings of the IEEE Conference on Computer Vision and Pattern Recognition Workshops}, pages 1186--1195, 2018.

\bibitem[\protect\citeauthoryear{Mees \bgroup \em et al.\egroup }{2016}]{27}
Oier Mees, Andreas Eitel, and Wolfram Burgard.
\newblock Choosing smartly: Adaptive multimodal fusion for object detection in changing environments.
\newblock In {\em 2016 IEEE/RSJ International Conference on Intelligent Robots and Systems (IROS)}, pages 151--156. IEEE, 2016.

\bibitem[\protect\citeauthoryear{Park \bgroup \em et al.\egroup }{2018}]{12}
Kihong Park, Seungryong Kim, and Kwanghoon Sohn.
\newblock Unified multi-spectral pedestrian detection based on probabilistic fusion networks.
\newblock {\em Pattern Recognition}, 80:143--155, 2018.

\bibitem[\protect\citeauthoryear{Perot \bgroup \em et al.\egroup }{2020}]{28}
Etienne Perot, Pierre De~Tournemire, Davide Nitti, Jonathan Masci, and Amos Sironi.
\newblock Learning to detect objects with a 1 megapixel event camera.
\newblock {\em Advances in Neural Information Processing Systems}, 33:16639--16652, 2020.

\bibitem[\protect\citeauthoryear{Redmon}{2016}]{5}
J~Redmon.
\newblock You only look once: Unified, real-time object detection.
\newblock In {\em Proceedings of the IEEE conference on computer vision and pattern recognition}, 2016.

\bibitem[\protect\citeauthoryear{Ren \bgroup \em et al.\egroup }{2016a}]{10}
Shaoqing Ren, Kaiming He, Ross Girshick, and Jian Sun.
\newblock Faster r-cnn: Towards real-time object detection with region proposal networks.
\newblock {\em IEEE transactions on pattern analysis and machine intelligence}, 39(6):1137--1149, 2016.

\bibitem[\protect\citeauthoryear{Ren \bgroup \em et al.\egroup }{2016b}]{15}
Shaoqing Ren, Kaiming He, Ross Girshick, and Jian Sun.
\newblock Faster r-cnn: Towards real-time object detection with region proposal networks.
\newblock {\em IEEE transactions on pattern analysis and machine intelligence}, 39(6):1137--1149, 2016.

\bibitem[\protect\citeauthoryear{Ross and Doll{\'a}r}{2017}]{1}
T-YLPG Ross and GKHP Doll{\'a}r.
\newblock Focal loss for dense object detection.
\newblock In {\em proceedings of the IEEE conference on computer vision and pattern recognition}, pages 2980--2988, 2017.

\bibitem[\protect\citeauthoryear{Takumi \bgroup \em et al.\egroup }{2017}]{20}
Karasawa Takumi, Kohei Watanabe, Qishen Ha, Antonio Tejero-De-Pablos, Yoshitaka Ushiku, and Tatsuya Harada.
\newblock Multispectral object detection for autonomous vehicles.
\newblock In {\em Proceedings of the on Thematic Workshops of ACM Multimedia 2017}, pages 35--43, 2017.

\bibitem[\protect\citeauthoryear{Tang \bgroup \em et al.\egroup }{2022}]{34}
Linfeng Tang, Jiteng Yuan, Hao Zhang, Xingyu Jiang, and Jiayi Ma.
\newblock Piafusion: A progressive infrared and visible image fusion network based on illumination aware.
\newblock {\em Information Fusion}, 83:79--92, 2022.

\bibitem[\protect\citeauthoryear{Wagner \bgroup \em et al.\egroup }{2016}]{8}
J{\"o}rg Wagner, Volker Fischer, Michael Herman, Sven Behnke, et~al.
\newblock Multispectral pedestrian detection using deep fusion convolutional neural networks.
\newblock In {\em ESANN}, volume 587, pages 509--514, 2016.

\bibitem[\protect\citeauthoryear{Wang \bgroup \em et al.\egroup }{2023}]{3}
Chien-Yao Wang, Alexey Bochkovskiy, and Hong-Yuan~Mark Liao.
\newblock Yolov7: Trainable bag-of-freebies sets new state-of-the-art for real-time object detectors.
\newblock In {\em 2023 IEEE/CVF Conference on Computer Vision and Pattern Recognition (CVPR)}, pages 7464--7475, 2023.

\bibitem[\protect\citeauthoryear{Wang \bgroup \em et al.\egroup }{2024}]{4}
Ao~Wang, Hui Chen, Lihao Liu, Kai CHEN, Zijia Lin, Jungong Han, and Guiguang Ding.
\newblock {YOLO}v10: Real-time end-to-end object detection.
\newblock In {\em The Thirty-eighth Annual Conference on Neural Information Processing Systems}, 2024.

\bibitem[\protect\citeauthoryear{Zhang \bgroup \em et al.\egroup }{2021}]{38}
Lu~Zhang, Zhiyong Liu, Xiangyu Zhu, Zhan Song, Xu~Yang, Zhen Lei, and Hong Qiao.
\newblock Weakly aligned feature fusion for multimodal object detection.
\newblock {\em IEEE Transactions on Neural Networks and Learning Systems}, 2(2):1--15, 2021.

\bibitem[\protect\citeauthoryear{Zhang \bgroup \em et al.\egroup }{2023}]{37}
Shilong Zhang, Xinjiang Wang, Jiaqi Wang, Jiangmiao Pang, Chengqi Lyu, Wenwei Zhang, Ping Luo, and Kai Chen.
\newblock Dense distinct query for end-to-end object detection.
\newblock In {\em Proceedings of the IEEE/CVF conference on computer vision and pattern recognition}, pages 7329--7338, 2023.

\bibitem[\protect\citeauthoryear{Zhang \bgroup \em et al.\egroup }{2024}]{zhang2024meddet}
Zeyu Zhang, Nengmin Yi, Shengbo Tan, Ying Cai, Yi~Yang, Lei Xu, Qingtai Li, Zhang Yi, Daji Ergu, and Yang Zhao.
\newblock Meddet: Generative adversarial distillation for efficient cervical disc herniation detection.
\newblock In {\em 2024 IEEE International Conference on Bioinformatics and Biomedicine (BIBM)}, pages 4024--4027. IEEE, 2024.

\end{thebibliography}
\end{document}